\documentclass[11pt]{article}

\usepackage[preprint]{acl}

\usepackage{times}
\usepackage{latexsym}

\usepackage[T1]{fontenc}

\usepackage[utf8]{inputenc}

\usepackage{microtype}

\usepackage{inconsolata}

\usepackage{graphicx}
\usepackage{booktabs}
\usepackage{multirow}
\usepackage{amssymb}
\usepackage{enumitem}
\usepackage{url}
\usepackage{hyperref}

\usepackage{xspace}

\newcommand{\nextfund}{\textsc{NextFund}\xspace}

\title{\nextfund :~A Unified Performance Tracking Platform \\ for Agentic Portfolio Management}

\author{%
  Changlun~Li\thanks{Correspondence:~tiger@paradoox.ai}
  \quad
  Peixian~Ma
  \quad
  Qiqi~Duan
  \quad
  Zhenyu~Lin
  \quad
  Peineng Wu
  \\[4pt]
  Paradoox AI Research
  \\[4pt]
}

\begin{document}
\maketitle

\begin{abstract}
Large language models (LLMs) based agents are beginning to participate in portfolio construction and market analysis, where decisions must be justified under evolving information and risk constraints. Current assessment practice, however, remains poorly aligned with this setting: many studies rely on static examinations or report only terminal portfolio returns, while the intermediate evidence, analyst judgments, and execution steps that produced those returns stay largely invisible. We introduce \nextfund, an evaluation platform that makes financial-agent behavior observable under live market conditions. The platform couples time-consistent market access, coordinated multi-agent analysis, and persistent logging of the full decision path from observation to trade. Through an interactive Trading Arena, users can compare models across markets, inspect equity curves, and drill from leaderboard outcomes down to individual justifications. We present \nextfund on Hong Kong, U.S., and China A-share equities, illustrating how inspectable decision histories enable fairer benchmarking and more actionable diagnosis. Our demo is available at \url{https://paradoox.cn/nextfund/}.
\end{abstract}

\section{Introduction}
Recent advancement of large language models (LLMs) has driven a shift from static financial text analysis toward agentic portfolio management, where LLMs-based agents can plan, gather evidence, and act under live market conditions~\citep{tradingagents,deepfund,li2025investorbench,yu2025finmem,xiong2025quantagent,aitrader,twinmarket}. Unlike traditional quantitative systems that follow fixed features and hard-coded rules, LLM-based agents can combine prices, news, calendars, and portfolio constraints through multi-step reasoning before placing an order. This change widens what financial automation can do, but it also raises a crucial tracking question: can an agent's portfolio decisions be shown to be reliable, comparable across models, and open to systematic improvement when markets move in real time?

Existing benchmarks answer this question only in part. Many focus on task coverage or final performance metrics such as cumulative return and Sharpe ratio, while giving little insight into how an action was formed~\citep{li2025investorbench,tradingagents,aitrader}. As a result, users who need to approve, monitor, or improve agent strategies still face three recurring problems:

\noindent\textbf{Incomplete Evaluation.} Standard quizzes and generic leaderboards rarely test whether an agent can work with live market information, follow portfolio constraints, and stay stable across different market conditions.

\noindent\textbf{Opaque Failure Diagnosis.} When an agent uses unsupported evidence, mishandles tools, or drifts from its investment mandate, developers often cannot tell whether the fault lies in retrieval, analysis, synthesis, or execution, and thus cannot improve the system with clear, repeatable feedback.

\noindent\textbf{Lost Evaluation Traces.} Decision logs, error cases, reviewer notes, and tool-call histories are often discarded after a run. Institutions therefore gain little reusable data for later prompt revision or model adaptation.

These problems grow more serious in multi-step workflows. An early error in evidence selection or time alignment can lead to a costly portfolio change. Backtests that look only at final profit and loss therefore miss the failure modes that matter most for risk control and review. What is needed is a unified performance tracking platform where agentic portfolio decisions can be measured at each step, compared across models, and retained for later improvement.

\begin{table*}[t]
    \centering
    \setlength{\tabcolsep}{6pt}
    \small
    \begin{tabular}{lcccc}
        \toprule
            \textbf{Method} & \textbf{Markets} & \textbf{Live} & \textbf{Trace} & \textbf{Arena} \\
        \midrule
            TradingAgents~\citep{tradingagents} & US & \checkmark & $\times$ & $\times$ \\
            FinMem~\citep{yu2025finmem} & US & $\times$ & $\times$ & $\times$ \\
            InvestorBench~\citep{li2025investorbench} & US, Crypto & $\times$ & $\times$ & $\times$ \\
            DeepFund~\citep{deepfund} & US & \checkmark & $\times$ & \checkmark \\
            TwinMarket~\citep{twinmarket} & CN & $\times$ & $\times$ & $\times$ \\
            QuantAgent~\citep{xiong2025quantagent} & US, Crypto, Comm. & $\times$ & Partial & $\times$ \\
            AI-Trader~\citep{aitrader} & US, CN, Crypto & \checkmark & $\times$ & Partial \\
        \midrule
            \textbf{\nextfund (Ours)} & \textbf{HK, US, CN} & \checkmark & \checkmark & \checkmark \\
        \bottomrule
    \end{tabular}
    \caption{Comparison of \nextfund with prior systems for agentic portfolio management. \textbf{Markets}: covered trading venues. \textbf{Live}: forward evaluation on live market streams. \textbf{Trace}: persistent end-to-end decision logs. \textbf{Arena}: interactive interface for leaderboard comparison and inspection.}
    \label{tab:comparison}
\end{table*}

We present \nextfund, a unified performance tracking platform for agentic portfolio management under these requirements. As shown in Table~\ref{tab:comparison}, earlier systems provide multi-agent trading, live testing, or interactive dashboards, yet rarely combine live multi-market tracking with reusable decision traces and arena-style inspection. \nextfund addresses this gap by recording observations, intermediate analyst outputs, and executed actions under a shared point-in-time view of the market. Its web-based Trading Arena enables leaderboard comparison and step-wise inspection, allowing users to move from performance differences to underlying agent behaviors.

Our contributions are as follows:
\begin{itemize}[leftmargin=*,itemsep=2pt,parsep=0pt,topsep=2pt]
    \item \textbf{A unified performance tracking platform.} \nextfund unifies multi-market data access, multi-agent coordination, and portfolio execution under common schemas and synchronized time, supporting fair comparison across models.
    \item \textbf{End-to-end decision tracing.} \nextfund systematically records observations, analyst signals, portfolio decisions, and execution outcomes, thereby transforming opaque agent operations into transparent performance histories.
    \item \textbf{An interactive Trading Arena.} The interface provides cross-market leaderboards, trajectory comparison, and trade-level rationale views, helping users move from summary scores to concrete diagnosis and improvement.
\end{itemize}

\section{Related Work}
\subsection{Financial Agentic Frameworks}
FinLLMs have moved financial AI from document understanding toward systems that retrieve evidence, reason over market state, and issue portfolio actions~\citep{wu2023bloomberggpt,yang2023fingpt,lee2024finllmsurvey,nie2024survey}. Building on this shift, recent agentic frameworks organize memory, personas, and specialist roles for trading and allocation~\citep{yu2025finmem,yu2024fincon,tradingagents,xiong2025quantagent,chen2025mentor,twinmarket}, while adjacent toolchains and RL environments supply brokerage interfaces or executable market simulators~\citep{qfinzero,finreporting,liu2020finrl,liu2022finrlmeta}. These systems advance agentic portfolio management, but they mainly optimize strategy construction or simulated returns; persistent, cross-model performance tracking under a shared live protocol remains secondary.

\subsection{Agent Tracking and Benchmarking}
Most financial LLM benchmarks still emphasize knowledge, reasoning, or trustworthiness rather than sequential portfolio decisions~\citep{xie2023pixiu,islam2023financebench,xie2024finben,liu2025findabench,peng2025multifinben,kang2026quanteval,hu-etal-2025-fintrust}. Decision-oriented suites move closer to deployment by testing agentic trading and allocation tasks~\citep{li2025investorbench,chen2026cnbuzz2portfolio,chen2026stockbenchcan}, and live protocols further reduce contamination by evaluating agents on post-cutoff market streams~\citep{deepfund,aitrader,yu2025livetradebenchseekingrealworldalpha}. Even so, many trackers privilege terminal P\&L, cover a narrow venue set, or retain only partial intermediate state. \nextfund targets this gap as a unified performance tracking platform that couples live multi-market evaluation with end-to-end decision traces and arena-style inspection.

\section{\nextfund System}
\label{sec:system}

\begin{figure*}[t]
    \centering
    \includegraphics[width=\textwidth]{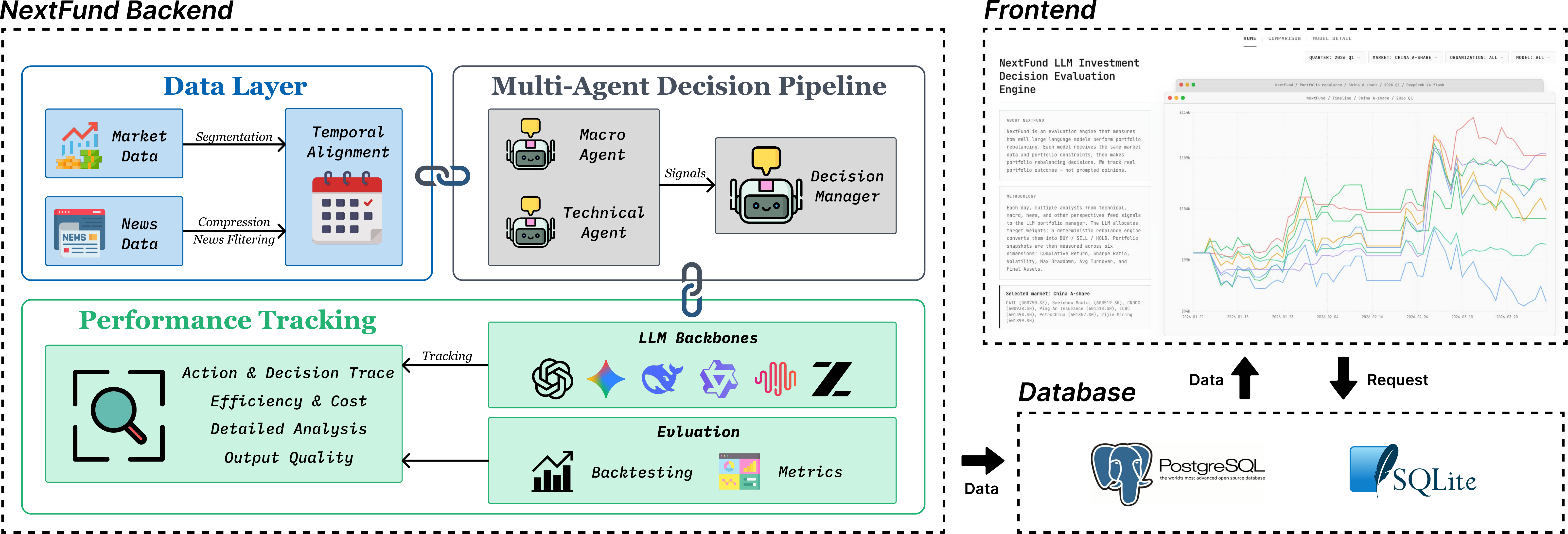}
    \caption{
        Overview of the \nextfund framework. The system integrates point-in-time data processing, multi-agent decision making, performance evaluation, and traceable inspection. The database module uses PostgreSQL for persistent backend data storage and SQLite for rapid I/O with the frontend.
    }
    \label{fig:arch}
\end{figure*}

\nextfund~serves as an evaluation layer connecting market data, multi-agent reasoning, and portfolio execution. Figure~\ref{fig:arch} summarizes the system architecture, which aligns information access, agent outputs, and decision traces.

\subsection{Data Layer}
\noindent
\nextfund begins with a data pipeline that turns heterogeneous market feeds into a shared, time-consistent observation set for agentic portfolio management. The pipeline covers price bars, macro snapshots, and hotlist news across U.S., China, and Hong Kong equities, then compresses narrative sources into a unified digest before any specialist agent is invoked.

\noindent\textbf{Market Data.}
For each candidate universe in the U.S., China, and Hong Kong markets, we collect daily equity price and volume bars and write them into a shared OHLCV store. From these bars, the pipeline derives market features used by later specialist analysts, while macro snapshots are retained alongside the price series for audit and recovery. All quantitative records are time-stamped so that downstream serving can enforce a consistent point-in-time view of the market.

\noindent\textbf{News Data.}
We collect time-stamped financial news as market-specific hotlists that feed the shared observation pipeline. 
%
For China A-shares and Hong Kong, the two markets share the same Chinese-language news sources, built from finance hot-rank articles on portals such as Sina Finance, East Money, and Tencent Finance. For the U.S., we retain English-language articles with publisher metadata from sources including Yahoo Finance and CNBC. Before compression, each article is cleaned for duplicates, aligned to a unified timestamp, and retained only if it passes basic quality filters.

\subsection{Multi-Agent Decision Pipeline}
\noindent
Time-aligned observations from the data layer are consumed by a multi-agent decision pipeline that maps market state to portfolio-level allocation proposals. The pipeline is staged into separate steps: specialized analysis is kept apart from constrained synthesis, and only proposals that satisfy execution constraints are forwarded to a shared evaluation layer. This separation prevents a single model from collapsing evidence gathering, risk control, and trade generation into one opaque step.

\noindent\textbf{Specialist Analysis.}
A set of dedicated analysts examines the shared decision state from complementary perspectives, including price dynamics, news narratives, macroeconomic conditions, and policy or risk cues. Each analyst returns a structured intermediate signal, thereby encoding directional judgment together with supporting justification. An upstream planner may activate the full analyst roster or a context-dependent subset; because activation and outputs are explicit, subsequent review can attribute each contribution to a specific specialist.

\noindent\textbf{Constrained Decision Synthesis.}
A decision manager aggregates the analyst signals subject to explicit execution constraints, including long-only exposure, position limits, cash availability, and transaction costs. The manager produces a textual rationale together with target allocations over the candidate universe and cash. By enforcing feasibility at this stage, the pipeline converts research judgments into portfolio proposals that remain consistent with operational and risk requirements.

\noindent\textbf{Handoff to Evaluation.}
The resulting proposals are transferred to evaluation layer, where they are instantiated as trades for backtesting and behavioral analysis. Maintaining boundary among research evidence, feasibility checks, and executed actions enables systematic attribution: performance shortfalls can be traced to deficiencies in analysis, position sizing, or execution, rather than remaining entangled in an end-to-end black box.

\subsection{Performance Tracking and Provenance}
\noindent
The preceding pipelines produce observations and allocation proposals; performance tracking is what makes those artifacts comparable, diagnosable, and reusable. In line with the platform goal stated in the title, \nextfund treats tracking not as an after-the-fact log dump, but as a first-class data plane of agentic portfolio management: every evaluation episode is retained as a structured provenance graph under a shared clock, so that terminal returns can be traced back to intermediate judgments and executed actions.

\noindent\textbf{Write-Through Integration.}
Tracking is embedded in the multi-agent workflow rather than attached after execution. When an analyst emits a signal, or when the decision manager commits a target allocation, the corresponding record is persisted immediately together with its justification and prompt context. Each trading day is anchored by a portfolio snapshot that serves as the join key for that day's signals, decisions, and news digests. Historical decisions and digests can also be read back into later runs as memory, closing the loop between recording and reasoning.

\noindent\textbf{From Traces to Improvement.}
Because traces are complete and time-aligned, the same substrate supports cross-model comparison, failure attribution, and iterative refinement. Reviewers can ask whether a rebalance followed coherent evidence aggregation or reflected a localized fault in retrieval, synthesis, or execution; retained trajectories further supply material for prompt revision and supervised adaptation. 
The interactive demo application in Section~\ref{sec:demo} is the user-facing interface over this tracking layer.

\section{Evaluation}
\label{sec:evaluation}

We assess \nextfund as an evaluation substrate for agentic portfolio management. The central question is whether the platform enables fair, time-consistent, and inspectable comparison of frontier LLM agents under a shared live protocol. 

\subsection{Task Setup}

All runs use the same multi-agent workflow, point-in-time data access, asset universe, and portfolio constraints are in Appendix~\ref{app:additional}.

\noindent\textbf{Evaluation Windows.}
Experiments cover the first two quarters of 2026 in full: 2026~Q1 ($2026$-$01$-$01$ to $2026$-$03$-$31$) and 2026~Q2 ($2026$-$04$-$01$ to $2026$-$06$-$30$). Within each window, agents rebalance on business days with synchronized prices, news digests, and account states.

\noindent\textbf{Markets and Universes.}
We evaluate three equity markets (U.S., China A-shares, and Hong Kong), each with a fixed seven-name universe spanning major large-cap names in that venue. Each run starts with a cash endowment of \$100{,}000 and follows long-only weight-based rebalancing under shared position and cash constraints.

\noindent\textbf{Models.}
We compare eight frontier LLMs as interchangeable backbones of the same agent stack, covering both proprietary and open-weight model families. 
Details are listed in Appendix~\ref{app:models-universes}.


\noindent\textbf{Metrics.}
We report cumulative return, Sharpe ratio, volatility, maximum drawdown, and turnover to capture performance, risk, and trading behavior in Section~\ref{sec:performance}.
Beyond aggregate metrics, \nextfund retains the analyst outputs, portfolio decisions, rationales, and execution records of each run, which support the tracable analysis in Section~\ref{sec:case_study} and the interactive inspection in Section~\ref{sec:demo}.

\begin{table*}[t]
    \centering
    \setlength{\tabcolsep}{2.6pt}
    \small
    \begin{tabular}{lrrrrrrrrrr}
        \toprule
        \multirow{2}{*}{\textbf{Model}} & \multicolumn{5}{c}{\textbf{2026 Q1}} & \multicolumn{5}{c}{\textbf{2026 Q2}} \\
        \cmidrule(lr){2-6} \cmidrule(lr){7-11}
         & Return (\%) & Sharpe & Vol (\%) & MDD (\%) & Turn (\%) & Return (\%) & Sharpe & Vol (\%) & MDD (\%) & Turn (\%) \\
        \midrule
        DeepSeek-V4-Flash & $-$4.51 & \textbf{$-$1.50} & 11.82 & 8.57 & 11.28 & 8.66 & 2.16 & 15.70 & 9.37 & 7.44 \\
        DeepSeek-V4-Pro & $-$11.86 & $-$2.94 & 16.69 & 15.36 & 8.15 & 10.27 & 2.16 & 18.65 & 11.86 & \textbf{3.32} \\
        Gemini-3.5-Flash & $-$9.61 & $-$4.45 & 9.00 & 11.62 & \textbf{7.81} & 3.28 & 0.86 & 16.36 & 11.13 & 6.45 \\
        GLM-5.1 & $-$5.48 & $-$3.65 & 6.13 & 7.11 & 11.57 & 0.18 & 0.12 & 14.83 & 12.00 & 12.57 \\
        GPT-5.4-Mini & $-$7.87 & $-$3.15 & 10.23 & 10.23 & 14.09 & 10.79 & \textbf{3.16} & \textbf{13.03} & \textbf{6.36} & 14.02 \\
        Kimi-K2.6 & \textbf{$-$2.89} & $-$2.30 & \textbf{5.04} & \textbf{4.39} & 8.91 & 8.95 & 2.18 & 16.07 & 9.84 & 5.96 \\
        MiniMax-M3 & $-$9.04 & $-$2.68 & 13.81 & 12.15 & 18.37 & 6.92 & 1.65 & 16.83 & 11.82 & 9.49 \\
        Qwen3.5-Flash & $-$11.15 & $-$2.78 & 16.51 & 15.22 & 15.43 & \textbf{12.23} & 2.77 & 16.93 & 9.63 & 10.58 \\
        \bottomrule
    \end{tabular}
    \caption{Live outcome metrics on U.S.\ equities under the shared \nextfund protocol. Return: cumulative return; Sharpe: Sharpe ratio; Vol: annualized volatility; MDD: max drawdown (absolute); Turn: avg turnover. Bold indicates best (higher is better for Return/Sharpe; lower for Vol/MDD/Turn). China/HK results are in Appendix~\ref{app:additional-results}.}
    \label{tab:live-us}
\end{table*}

\subsection{Cross-Model Performance Comparison}
\label{sec:performance}

Table~\ref{tab:live-us} compares the eight model backbones on the U.S.\ market across the two evaluation quarters. Results for China A-shares and Hong Kong equities are in Appendix~\ref{app:additional-results}.
The comparison reveals three main patterns. First, ranking depends on both the metric and the regime. In 2026~Q1, every backbone loses money in the U.S., but Kimi-K2.6 is the least negative on return ($-2.89\%$) and also records the best volatility ($5.04\%$) and drawdown ($4.39\%$), whereas DeepSeek-V4-Flash has the least negative Sharpe. In Q2 the book rebounds: Qwen3.5-Flash leads return ($12.23\%$), while GPT-5.4-Mini leads Sharpe ($3.16$), volatility ($13.03\%$), and drawdown ($6.36\%$); Gemini-3.5-Flash records the lowest turnover in Q1 ($7.81\%$) and DeepSeek-V4-Pro in Q2 ($3.32\%$). Second, return, risk, and trading intensity do not move together: a higher-return model is not automatically the most efficient, the most stable, or the least active, which is why \nextfund exposes the full metric vector rather than a single P\&L score. Third, the same protocol yields qualitatively different leaderboards in China and Hong Kong (Appendix~\ref{app:additional-results}), confirming that cross-market tracking is necessary for fair comparison.

\subsection{Case Study}
\label{sec:case_study}
%
To illustrate \nextfund's diagnostic capability, we compare two LLM backbones under the same market, period, asset universe, and input stream.\footnote{Interactive comparison is available at \url{https://paradoox.cn/nextfund/comparison}.}
The aligned traces connect portfolio trajectories with specialist signals and portfolio-manager actions, enabling inspection of where cross-model divergence emerges. 
We use the U.S.\ 2026~Q1 case, comparing Kimi-K2.6 and Qwen3.5-Flash, the highest- and lowest-return models in Table~\ref{tab:live-us} for that period.

\noindent\textbf{Signals can agree while decisions diverge.}
On sampled trading days in U.S.\ 2026~Q1, comparable analyst signal pairs (analyst $\times$ ticker) between Kimi-K2.6 and Qwen3.5-Flash agree on about $92.9\%$ of cases, suggesting that specialist views are often similar when both models see the same prices and news digests. Final BUY/SELL/HOLD actions, however, agree on only $36.4\%$ of ticker-days over the full 64-day window, and on $47$ days more than half of the seven tickers disagree. The action mixes explain much of the gap: Kimi-K2.6 remains conservative ($396$ HOLD / $29$ BUY / $23$ SELL), whereas Qwen3.5-Flash trades far more actively ($152$ HOLD / $160$ BUY / $136$ SELL). The comparison therefore surfaces a concrete failure mode for opaque evaluation: similar intermediate evidence need not imply similar portfolio behavior.

\noindent\textbf{Metric profiles separate return from risk and trading intensity.}
The same pair also illustrates why a single P\&L number is insufficient. In U.S.\ 2026~Q1, Kimi-K2.6 records the best return ($-2.89\%$), volatility ($5.04\%$), and drawdown ($4.39\%$) among the eight backbones, with moderate turnover ($8.91\%$). Qwen3.5-Flash finishes near the bottom on return ($-11.15\%$) while posting higher volatility ($16.51\%$), deeper drawdown ($15.22\%$), and nearly double the turnover ($15.43\%$). In Q2 the ranking flips on return, where Qwen3.5-Flash leads at $12.23\%$, yet GPT-5.4-Mini remains preferable on Sharpe, volatility, and drawdown. Head-to-head tracking thus shows that model differences are multi-dimensional. Aggressiveness in the decision layer can raise turnover and risk even when analyst signals look aligned, and the best return model in one quarter need not dominate risk-adjusted metrics in the next.

\section{Demo Application}
\label{sec:demo}

\begin{figure*}[t]
    \centering
    \includegraphics[width=\textwidth]{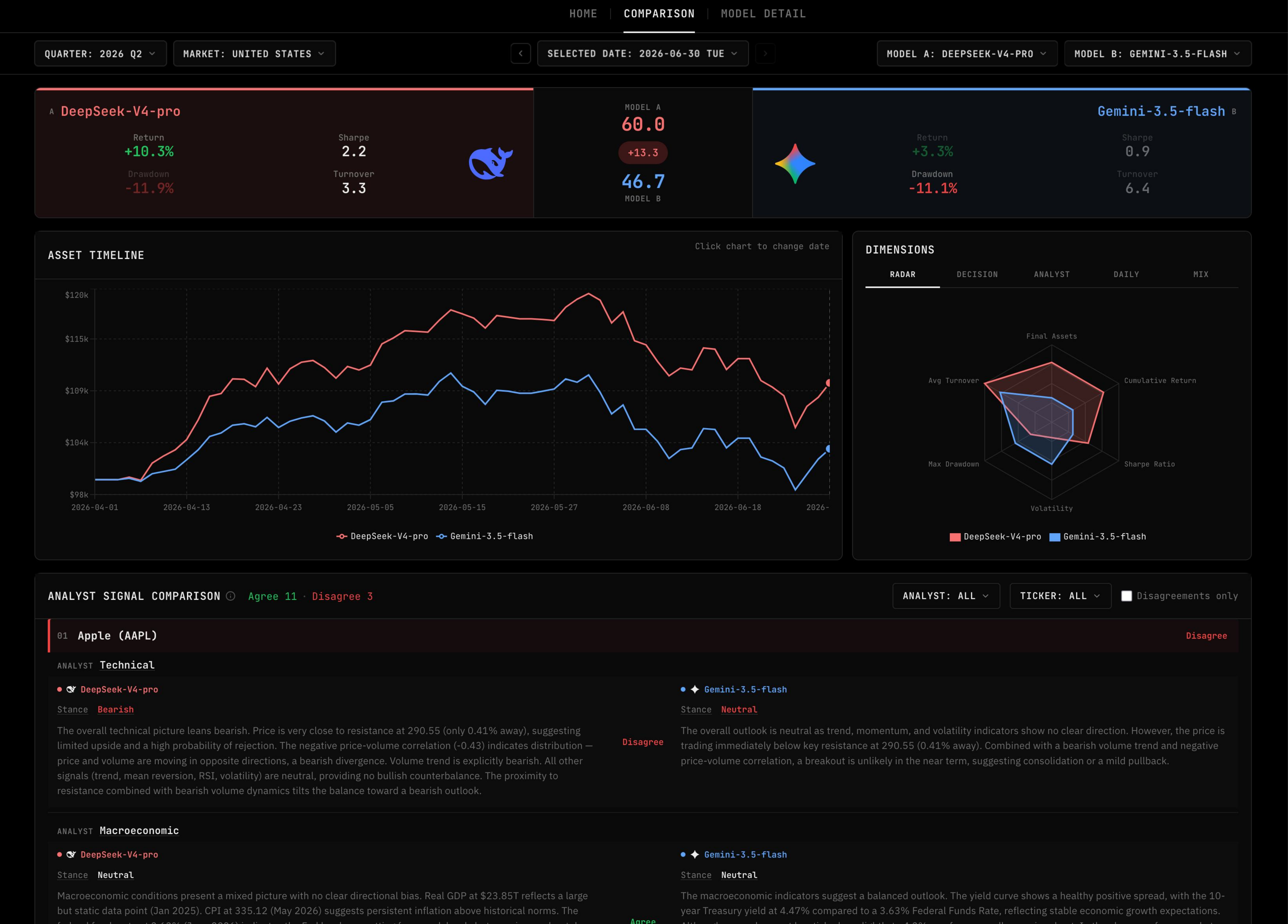}
    \caption{The interface of \nextfund demo. This interface enables users to select different LLM backbones, evaluation periods and markets. Users can also select a specific model to view detailed information and analyze the decision trajectory.}
    \label{fig:arena}
    \vspace{-0.2em}
\end{figure*}

Unlike benchmarks that mainly report end-state returns~\citep{li2025investorbench,tradingagents}, \nextfund connects portfolio outcomes with the decision traces behind them. The demo supports a two-stage workflow for researchers and practitioners: identifying agent behavioral differences (Section~\ref{sec:arena}) and tracing them back to intermediate decisions (Section~\ref{sec:tracking}).

\subsection{Workflow Overview}
Users first select a market among China, U.S., and Hong Kong and an evaluation window among 2026~Q1 and 2026~Q2. 
Given this selection, the interface loads the corresponding runs under the same universe, cash endowment, and rebalancing rules used in Section~\ref{sec:evaluation}. 
From there, users can move to a comparison view for side-by-side model ranking and trajectory inspection, or to a detail view for step-wise decision tracking. 

\subsection{Model Arena}
\label{sec:arena}
The Model Arena provides the entry point for cross-model analysis. 
Users select a market, evaluation period, and set of model runs to compare. The interface presents cumulative return, Sharpe ratio, volatility, maximum drawdown, and turnover together with equity curves, allowing users to identify when and how portfolio trajectories diverge under identical evaluation conditions.
Because each run is linked to its retained execution trace, users can directly move from a performance difference to the corresponding decision records. For example, a model with higher return but higher turnover can be selected for further inspection to understand whether the difference arises from more frequent actions or different portfolio allocation choices. 

\subsection{Detail Tracking}
\label{sec:tracking}
The Detail Tracking view enables users to inspect an individual run at the decision level. For each trading day and ticker, the interface links specialist outputs, portfolio-manager decisions, executed actions, and textual rationales along a shared timeline. Users can examine whether two agents diverge during upstream analysis, decision synthesis, or execution. For instance, after observing a trajectory divergence in the Model Arena, users can select the corresponding period and compare analyst signals with final actions to determine whether similar evidence led to different portfolio decisions. This workflow transforms aggregate performance comparison into a traceable diagnosis: rather than only identifying which model performed differently, \nextfund helps users inspect how and where the difference emerged.

\section{Conclusion and Future Work}
We present \nextfund, a unified live platform for agentic portfolio management.
By combining multi-market data access, multi-agent analysis, and end-to-end decision tracing, \nextfund addresses the three recurring problems identified above: incomplete evaluation, opaque failure diagnosis, and lost evaluation traces. The demo application further turns recorded trajectories into practical tools for ranking, inspection, and iterative improvement. In future work, we plan to cover more asset classes, add stronger checks for unsupported claims and risk compliance, and reuse retained trajectories for prompt optimization and agentic learning.

\section*{Limitations}
The current release centers on equity rebalancing across three markets with a fixed analyst roster. Support for derivatives, multi-currency hedging, and richer compliance rules is left to future work. Even with synchronized clocks and structured logs, textual rationales may remain incomplete or only loosely aligned with latent model computation; human review is still required for consequential decisions. Live evaluation also depends on third-party market feeds whose licenses may prohibit redistributing raw data; public artifacts therefore emphasize schemas, traces, and interfaces rather than proprietary market dumps.

\section*{Ethical Considerations}
Financial agent evaluation can influence high-stakes judgments. \nextfund is intended as an auditing and comparison aid, not as an investment advisor. Demonstration outcomes should not be the sole basis for investment or regulatory decisions without independent checks against primary sources. Users should examine provenance, decision histories, and risk metrics, and interpret leaderboard results as comparative evidence under a stated protocol rather than as forecasts of future performance.

\bibliography{reference}

\clearpage
\appendix
\section{Additional Details}
\label{app:additional}
\subsection{Detailed Database Schema}

\noindent\textbf{Tracking Trajectory Schema.}
\nextfund persists each evaluation episode as a time-aligned trajectory that can be reconstructed from the live tracking API.
Table~\ref{tab:trace-schema} summarizes the core fields of a per-ticker decision record returned by the trajectory endpoint, including the attached analyst signals used for diagnosis.

\begin{table}[h!]
\centering
\setlength{\tabcolsep}{3pt}
\footnotesize
\begin{tabular}{@{}p{0.32\columnwidth} p{0.10\columnwidth} p{0.48\columnwidth}@{}}
\toprule
\textbf{Field} & \textbf{Type} & \textbf{Description} \\
\midrule
\texttt{trading\_date} & string & Business date of the decision record. \\
\texttt{portfolio\_id} & string & Identifier linking the record to the same-day portfolio snapshot. \\
\texttt{ticker} & string & Asset symbol under consideration. \\
\texttt{action} & string & Executed action (\texttt{Buy}, \texttt{Sell}, or \texttt{Hold}). \\
\texttt{shares} & number & Number of shares traded; zero for hold actions. \\
\texttt{price} & number & Reference or execution price used by the record. \\
\texttt{justification} & string & Textual rationale for the executed action. \\
\texttt{linked\_trace\_id} & string & Provenance identifier for the decision trace. \\
\texttt{workflow\_decision} & object & Portfolio-manager workflow metadata, including selected analysts. \\
\texttt{analyst\_signals} & array & Specialist outputs attached to the decision. \\
\bottomrule
\end{tabular}
\caption{Schema of a tracking trajectory decision record in \nextfund.}
\label{tab:trace-schema}
\end{table}

\noindent\textbf{Price Data Schema.}
Market prices are stored as daily OHLCV bars under a shared point-in-time protocol.
Table~\ref{tab:price-schema} lists the fields of the equity price table used by the data layer.

\begin{table}[h]
\centering
\setlength{\tabcolsep}{3pt}
\footnotesize
\begin{tabular}{@{}p{0.28\columnwidth} p{0.12\columnwidth} p{0.50\columnwidth}@{}}
\toprule
\textbf{Field} & \textbf{Type} & \textbf{Description} \\
\midrule
\texttt{pool\_name} & text & Equity pool name (e.g., \texttt{cn}, \texttt{us}, \texttt{hk}). \\
\texttt{market} & text & Market identifier of the listed venue. \\
\texttt{symbol} & text & Ticker symbol in the evaluation universe. \\
\texttt{trade\_date} & text & Trading date in \texttt{YYYY-MM-DD} format. \\
\texttt{open} & real & Opening price. \\
\texttt{high} & real & Highest price. \\
\texttt{low} & real & Lowest price. \\
\texttt{close} & real & Closing price. \\
\texttt{volume} & integer & Trading volume. \\
\texttt{source} & text & Upstream data provider identifier. \\
\bottomrule
\end{tabular}
\caption{Schema of the daily equity price table in \nextfund.}
\label{tab:price-schema}
\end{table}

\begin{table}[t!]
\centering
\setlength{\tabcolsep}{3pt}
\footnotesize
\begin{tabular}{@{}p{0.28\columnwidth} p{0.12\columnwidth} p{0.50\columnwidth}@{}}
\toprule
\textbf{Field} & \textbf{Type} & \textbf{Description} \\
\midrule
\texttt{id} & integer & Auto-increment primary key. \\
\texttt{source} & text & News source identifier (e.g., \texttt{sina}). \\
\texttt{title} & text & Article headline. \\
\texttt{url} & text & Canonical article URL. \\
\texttt{publish\_time} & text & Publication timestamp. \\
\texttt{crawl\_time} & text & Timestamp when the article was crawled. \\
\texttt{crawl\_date} & text & Calendar date used for day-level alignment. \\
\texttt{media} & text & Publisher or media outlet name. \\
\texttt{top\_num} & integer & Hotlist rank or popularity index. \\
\texttt{category} & text & Source category or channel tag. \\
\texttt{content} & text & Article body text when available. \\
\texttt{author} & text & Author byline when available. \\
\texttt{keywords} & text & Extracted or provided keywords. \\
\texttt{word\_count} & integer & Approximate body length in words or characters. \\
\texttt{related\_stocks} & text & Linked tickers associated with the article. \\
\texttt{created\_at} & timestamp & Database insertion time. \\
\bottomrule
\end{tabular}
\caption{Schema of the hotlist news table in \nextfund.}
\label{tab:news-schema}
\end{table}

\noindent\textbf{News Data Schema.}
News items are collected from market-specific hotlists and aligned to the evaluation clock before compression into digests.
Table~\ref{tab:news-schema} summarizes the core fields of the news table.

\section{Additional Evaluations}

\subsection{Evaluation Models and Universes}
\label{app:models-universes}

\begin{table*}[t!]
    \centering
    \setlength{\tabcolsep}{2.6pt}
    \small
    \begin{tabular}{lrrrrrrrrrr}
        \toprule
        \multirow{2}{*}{\textbf{Model}} & \multicolumn{5}{c}{\textbf{2026 Q1}} & \multicolumn{5}{c}{\textbf{2026 Q2}} \\
        \cmidrule(lr){2-6} \cmidrule(lr){7-11}
         & Return (\%) & Sharpe & Vol (\%) & MDD (\%) & Turn (\%) & Return (\%) & Sharpe & Vol (\%) & MDD (\%) & Turn (\%) \\
        \midrule
        DeepSeek-V4-Flash & 3.36 & 0.86 & 17.11 & 3.86 & 18.55 & $-$14.98 & $-$4.32 & 14.52 & 15.64 & 16.54 \\
        DeepSeek-V4-Pro & 7.29 & 1.51 & 19.86 & 4.32 & 8.00 & $-$15.31 & $-$4.83 & 13.36 & 15.75 & 10.40 \\
        Gemini-3.5-Flash & 4.18 & 0.82 & 23.27 & 7.52 & \textbf{7.49} & $-$12.38 & $-$5.11 & 10.06 & 12.38 & 12.50 \\
        GLM-5.1 & \textbf{9.76} & 2.20 & 17.67 & 3.70 & 10.20 & $-$12.58 & $-$4.63 & 11.29 & 12.75 & \textbf{9.49} \\
        GPT-5.4-Mini & 0.85 & 0.42 & \textbf{9.13} & \textbf{2.79} & 16.84 & $-$11.97 & $-$5.65 & 8.82 & 12.20 & 17.80 \\
        Kimi-K2.6 & 9.52 & \textbf{2.42} & 15.51 & 3.31 & 13.01 & \textbf{$-$7.66} & \textbf{$-$3.89} & \textbf{7.98} & \textbf{7.85} & 12.03 \\
        MiniMax-M3 & 7.02 & 1.66 & 17.18 & 4.08 & 22.09 & $-$17.21 & $-$5.57 & 13.17 & 18.70 & 18.72 \\
        Qwen3.5-Flash & $-$4.81 & $-$0.96 & 18.76 & 7.39 & 13.89 & $-$14.00 & $-$5.07 & 11.58 & 14.00 & 16.31 \\
        \bottomrule
    \end{tabular}
    \caption{Live outcome metrics on China A-share equities under the shared \nextfund protocol. Metric definitions follow Table~\ref{tab:live-us}.}
    \label{tab:live-cn}
\end{table*}

\begin{table*}[t!]
    \centering
    \setlength{\tabcolsep}{2.6pt}
    \small
    \begin{tabular}{lrrrrrrrrrr}
        \toprule
        \multirow{2}{*}{\textbf{Model}} & \multicolumn{5}{c}{\textbf{2026 Q1}} & \multicolumn{5}{c}{\textbf{2026 Q2}} \\
        \cmidrule(lr){2-6} \cmidrule(lr){7-11}
         & Return (\%) & Sharpe & Vol (\%) & MDD (\%) & Turn (\%) & Return (\%) & Sharpe & Vol (\%) & MDD (\%) & Turn (\%) \\
        \midrule
        DeepSeek-V4-Flash & 1.77 & 0.37 & 36.81 & 15.33 & 13.85 & 11.95 & 1.65 & 29.48 & 9.48 & 16.42 \\
        DeepSeek-V4-Pro & 14.63 & 1.63 & 37.66 & 11.01 & 9.10 & 1.34 & 0.35 & 22.07 & 8.46 & 7.92 \\
        Gemini-3.5-Flash & 18.56 & 1.86 & 40.91 & 13.03 & 8.45 & \textbf{23.41} & \textbf{2.87} & 30.48 & 7.98 & \textbf{5.78} \\
        GLM-5.1 & 7.02 & 1.46 & \textbf{19.98} & 7.59 & \textbf{5.50} & 0.47 & 0.19 & 20.60 & 8.92 & 7.09 \\
        GPT-5.4-Mini & 7.27 & 1.32 & 23.33 & 9.50 & 16.36 & 4.43 & 1.21 & \textbf{15.07} & \textbf{5.77} & 13.79 \\
        Kimi-K2.6 & \textbf{30.94} & \textbf{3.21} & 35.51 & \textbf{6.45} & 11.33 & 7.01 & 1.55 & 18.33 & 6.94 & 12.26 \\
        MiniMax-M3 & 10.24 & 1.29 & 34.99 & 11.23 & 19.63 & 8.79 & 1.43 & 25.35 & 8.83 & 15.27 \\
        Qwen3.5-Flash & 20.10 & 2.05 & 39.56 & 8.67 & 15.04 & 8.71 & 1.30 & 28.18 & 9.77 & 14.16 \\
        \bottomrule
    \end{tabular}
    \caption{Live outcome metrics on Hong Kong equities under the shared \nextfund protocol. Metric definitions follow Table~\ref{tab:live-us}.}
    \label{tab:live-hk}
\end{table*}

\noindent\textbf{Models.} We evaluate eight frontier LLMs as interchangeable backbones of the same multi-agent stack: DeepSeek-V4-Flash~\cite{deepseek2026v4}, DeepSeek-V4-Pro~\cite{deepseek2026v4}, Gemini-3.5-Flash~\cite{google2026gemini3.5flash}, GLM-5.1~\cite{zeng2026glm}, GPT-5.4-Mini~\cite{openai2026gpt54}, Kimi-K2.6~\cite{kimi_k2_2026}, MiniMax-M3~\cite{minimax_m3_2026}, and Qwen3.5-Flash~\cite{alibaba2026qwen35flash}. The set covers both proprietary and open-weight model families. All backbones share the same point-in-time data access, analyst roster, portfolio constraints, and logging schema.

\noindent\textbf{Asset Universes.}
Each market uses a fixed seven-name large-cap universe: the U.S.\ pool comprises AAPL, MSFT, NVDA, TSLA, AMZN, GOOGL, and JPM; the China A-share pool comprises 600519.SH, 300750.SZ, 601318.SH, 601899.SH, 601398.SH, 601857.SH, and 600938.SH; and the Hong Kong pool comprises 9992.HK, 0700.HK, 1810.HK, 2513.HK, 03690.HK, 0005.HK, and 1299.HK.

\subsection{Additional Market Results}
\label{app:additional-results}
This subsection reports the same outcome metrics as Table~\ref{tab:live-us} for China A-shares and Hong Kong equities. In China (Table~\ref{tab:live-cn}), GLM-5.1 leads return in 2026~Q1 ($9.76\%$), while Kimi-K2.6 leads Sharpe ($2.42$) and GPT-5.4-Mini leads volatility and drawdown; Gemini-3.5-Flash records the lowest turnover in Q1 ($7.49\%$). In Q2 all models are negative, and Kimi-K2.6 remains strongest across return, Sharpe, volatility, and drawdown, while GLM-5.1 has the lowest turnover ($9.49\%$). In Hong Kong (Table~\ref{tab:live-hk}), Kimi-K2.6 leads return ($30.94\%$), Sharpe ($3.21$), and drawdown ($6.45\%$) in Q1, with GLM-5.1 recording the lowest volatility and turnover; in Q2, Gemini-3.5-Flash leads return ($23.41\%$), Sharpe ($2.87$), and turnover ($5.78\%$), whereas GPT-5.4-Mini leads volatility and drawdown.

\subsection{License}
The \nextfund software and associated artifacts released with this paper are licensed under the Apache License~2.0.\footnote{\url{https://www.apache.org/licenses/LICENSE-2.0}}

\end{document}